# An evaluation of an algorithm for inductive learning of Bayesian belief networks using simulated data sets

**Constantin F. Aliferis and Gregory F. Cooper**
Section of Medical Informatics & Intelligent Systems Program,
University of Pittsburgh, B50A Lothrop Hall, 190 Lothrop St.
Pittsburgh, PA 15261

## Abstract

Bayesian learning of belief networks (BLN) is a method for automatically constructing belief networks (BNs) from data using search and Bayesian scoring techniques. K2 is a particular instantiation of the method that implements a greedy search strategy. To evaluate the accuracy of K2, we randomly generated a number of BNs and for each of those we simulated data sets. K2 was then used to induce the generating BNs from the simulated data. We examine the performance of the program, and the factors that influence it. We also present a simple BN model, developed from our results, which predicts the accuracy of K2, when given various characteristics of the data set.

## 1 INTRODUCTION

Bayesian belief networks (BNs) constitute a method for graphical representation of knowledge, based on explicitly defining probabilistic dependencies and independences among variables. A BN consists of a directed acyclic graph (DAG) that captures the dependencies and independences among nodes (corresponding to variables) of the graph, and a set of functions that give each variable's probability distribution, conditioned on the values of its parent nodes [Pearl 1988, Neapolitan 1990]. BNs are a state-of-the-art formal method for probabilistic modelling in decision-support systems [Cooper 1989].

Although BNs can reduce dramatically the number of probabilities that must be specified for a particular modelling task, relative to methods that do not exploit the independence relations among the domain variables, the knowledge acquisition (KA) problem is still challenging. To cope with the KA "bottleneck", researchers within the symbolic Artificial Intelligence (AI) Machine-learning (ML) community have developed methods for learning representations of knowledge automatically from collections of data sets [Shavlick 1990]. In the same spirit, researchers in the BN field have developed techniques which when given a set of variable observations, will try to find the BN (or depending on the method, the class of BNs) that most probably produced the data set (i.e., that best captures the variables relationships) [Cooper 1992, Pearl 1993, Fung 1990, Lam 1993, Singh 1993, Spirtes 1992, Suzuki 1993].

The pursuit of ML methods for BN construction is further motivated by the following applications areas: (a) exploratory statistical analysis, (b) comparison, confirmation, and discovery of scientific hypotheses, (c) partial substitution of classical multivariate analytic techniques [Cooper 1992, Aliferis 1993].

One method for BN ML is the Bayesian learning of BNs (BLN) method [Cooper 1992]. This method, when given a database of observations, searches a space of BNs, and scores them using a Bayesian scoring function. A particular instantiation of the method is the algorithm K2, which uses greedy search as the search strategy. K2 also requires as an input an ordering of the variables, such that no variable later in the ordering can "cause" (be the parent of) a variable earlier in the ordering [Cooper 1992]. It is assumed that temporal precedence and domain knowledge suffice for the determination of such an ordering. In the conclusions section we discuss methods for relaxing this assumption.

The goal of the research reported in the current paper is to investigate the accuracy of K2, and to identify data attributes that possibly determine its accuracy using simulated data as contrasted to real data. The problem with real data is that frequently a gold standard (i.e., the underlying BN process) is not known. Thus in such cases researchers measure how well the ML method models the domain indirectly, by measuring the predictive accuracy of the produced model. For an initial evaluation of K2 using this method, see [Herskovits 1991].

Using simulated data produced by well-specified models (gold-standard models) on the other hand enables us to overcome these difficulties and measure directly how well the ML method learns the model structure. An admitted limitation, however, is that the simulated data may not necessarily resemble closely the type of data we would obtain from samples in the real world. In a preliminary evaluation of K2 using this method, Cooper and Herskovits used simulated data from the ALARM network (a BN with 37 nodes and 46 arcs, developed to model the anesthesia emergencies in the operating room [Beinlich 1989]), to examine the number of arcs correctly found, and erroneously added by K2, given database sizes ranging from 100 to 10000 cases [Cooper 1992]. In this paper we describe experiments that extend the use of simulation to obtain insight into BN learning methods. In particular we describe the use of simulated data from a variety of BNs and we discuss not only the findings, but also which specific data attributes determine the accuracy of the algorithm, and how a user can infer the expected accuracy for a particular learning task.

## 2 METHODS

The experiment consists of the following parts: a) Generation of a set of BNs, which we call the *Gold standard BNs* (BNs-GS). For each belief network the number of variables was chosen randomly from the following values: 2, 10, 20, 30, 40, 50. The number of arcs was chosen randomly (i.e., a uniform probability distribution was used), so that between 0 and 10 arcs would point to any particular node. The ordinality of the variables (i.e., total number of possible values) was randomly chosen to be either two or three for all variables in a generated BN. After the structures were constructed, they were parameterized (i.e., conditional probabilities functions were determined for each node) randomly for each prior and conditional probability.
b) The set of generated BNs was given to the case generator. For each BN, the case generator constructed a set of simulated data using logic sampling [Henrion 1988]. The number of cases per BN was chosen randomly between 0 and 2000 cases.
c) The sets of simulated cases were given to K2, which constructed for each data set a BN. K2 had access to the correct ordering of the variables for each BN-GS. We will call the set of BNs produced by K2 the *Induced BNs* (BNs-I).
d) Finally, the sets of gold-standard BNs and the induced BNs were compared by a statistics module, which estimated descriptive statistics and the following two metrics for each BN-GS and BN-I pair: percentage of arcs in BNs-GS that are present in BNs-I (metric M1), and ratio of number of arcs in BNs-I that are absent in BNs-GS to the number of arcs in the corresponding BN-GS (metric M2). Additional analyses were performed on this output using a statistical package and appropriate techniques [Norusis 1992]. The diagram in Figure 1 depicts the experimental design.

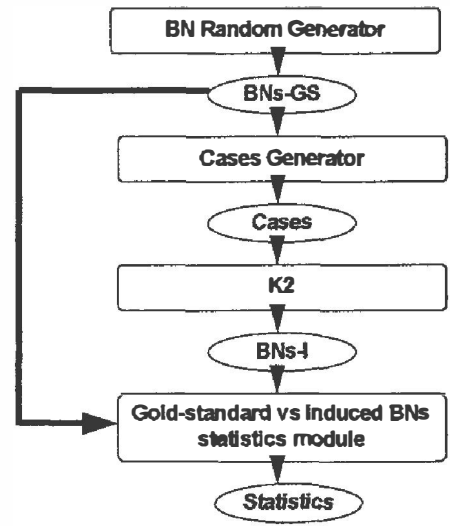

Figure 1. Flowchart of the Experimental Design

The experiment was performed using an integrated software package supporting belief network inference, learning, and simulated BN and case generation and evaluation, which we have developed. The software is written in Pascal and runs on an IBM RS 6000 workstation. For pragmatic reasons we decided to run the program in batch mode and analyze the results which were produced. We additionally developed using K2, a model of K2's accuracy (conditioned upon data attributes) for the purposes of empirical accuracy prediction.

## 3 RESULTS

A total of 67 BN pairs were generated and analyzed. Table 1 presents the descriptive statistics for the data attributes (number of variables, ordinality of variables, number of arcs, number of cases). To facilitate analysis, we additionally discretized the number of arcs and cases. Table 2 shows the corresponding information. Tables 3 and 4 present the descriptive statistics for the evaluation metrics we used, both in their original and discretized forms. As it is evident from Table 4, K2 discovered at least 70% of the arcs 94% of the time. In 94% of the cases, K2 did not add more than 10% arcs of the BN-GS arcs. The mean percentage of correctly found arcs (M1) is 91.6% and the mean ratio of superfluous arcs (M2) is 4.7%.

10   Aliferis and Cooper

Table 1: Descriptive Statistics for Data Attributes of BNs-GS

| variable | value | frequency % |
|---|---|---|
| number of variables | 2 | 6.0 |
| | 10 | 16.4 |
| | 20 | 26.9 |
| | 30 | 22.4 |
| | 40 | 14.9 |
| | 50 | 13.4 |
| ordinality of variables | 2 | 46.3 |
| | 3 | 53.7 |

| variable | mean | s.d. |
|---|---|---|
| number of arcs | 60.93 | 36.77 |
| number of cases | 1085.49 | 544.97 |

Table 2: Descriptive Statistics for Discretized Data Attributes

| | | frequency distribution % |
|---|---|---|
| number of arcs | 0-20 | 16.4 |
| | 21-60 | 37.3 |
| | 61-100 | 25.4 |
| | >100 | 20.9 |
| number of cases | 0-200 | 3.0 |
| | 201-500 | 17.9 |
| | 501-1000 | 22.4 |
| | 1001-1500 | 32.8 |
| | >1500 | 23.9 |

Table 3: Descriptive Statistics for Evaluation Metrics

| | mean | s.d. |
|---|---|---|
| M1 (%) | 91.6 | 11.7 |
| M2 (%) | 4.7 | 7.6 |

Table 4: Descriptive Statistics for Discretized Evaluation Metrics

| | value | frequency distribution % |
|---|---|---|
| M1 | 0-50 % | 1.5 |
| | 51-70% | 4.5 |
| | 71-90% | 28.4 |
| | 91-95% | 11.9 |
| | 96-98% | 13.4 |
| | >98% | 40.3 |
| M2 | 0-2% | 47.8 |
| | 3-5% | 19.4 |
| | 6-10% | 26.9 |
| | 11-30% | 4.5 |
| | 31-50% | 1.5 |
| | >50% | 0 |

We also analyzed the factors that influence the performance of K2. The nature of the data is such that the influences of the independent variables (number of variables, number of arcs, number of cases and variable ordinality) on the dependent ones (i.e., M1, M2), can not be analyzed appropriately with a linear model. Although we tried a number of variable transformations on the variables, an analysis of variance/covariance or multiple regression model was not applicable, due to violation of assumptions. Thus we applied a graphical analysis of the response surface, followed by fitting a non-linear regression model to the relationships that were revealed by the analysis.

Graphs I and II show the relationship between number of arcs, number of variables and number of cases for the case where ordinality is 2 or 3 (graphs I & II respectively). As we would expect from our design, the number of variables is uniformly distributed across the number of cases. For each number of variables, there is small variation of the corresponding arcs number (since we constrained the incoming arcs per variable in the generation process - as described in the methods section). Finally, the same observations hold true when ordinality is 3, although the spread of data points is somewhat more constrained. These graphs imply that we can eliminate the number of arcs from further consideration, since it is determined by the number of variables. Also they suggest that we might want to apply two different analyses, one for cases where variables were binary and one where they were ternary due to the somewhat different spread of data points.

Graph III shows the relationships between M1 & M2 and number of cases for the complete data set (i.e., both cases containing variables with ordinality 2 and ordinality 3). Similar relationships exist for the subset with ordinality 2 and the subset with ordinality 3. Graph IV shows the relationships between M1 & M2 and number of variables for the complete data set. Again similar plots have been produced (not shown here) for the subset with ordinality 2 and the subset with ordinality 3.

The graphs shown here support the following: (a) M1 appears to be asymptotically approaching 100% as cases increase (graph III),
(b) M2 appears to be asymptotically approaching 0 as cases increase (graph III),
(c) there is no clear form of covariation of M1, M2 and number of variables (graph IV).
In addition, even though for both binary and ternary variables the same nature (i.e. functional form) of covariation exists between M1 & M2 and cases, the function parameters should be assessed individually since the relevant plots (not shown here) have different spread of data points.

and for M2 is 0.8, when we model these metrics separately for ordinality of 2 and 3), and thus these GRAPH II. Relationship between arcs and cases when ordinality is 3. Data points corresponding to BNs with different numbers of variables are separated into 6 different groups. Numbers for each group denote number of variables.

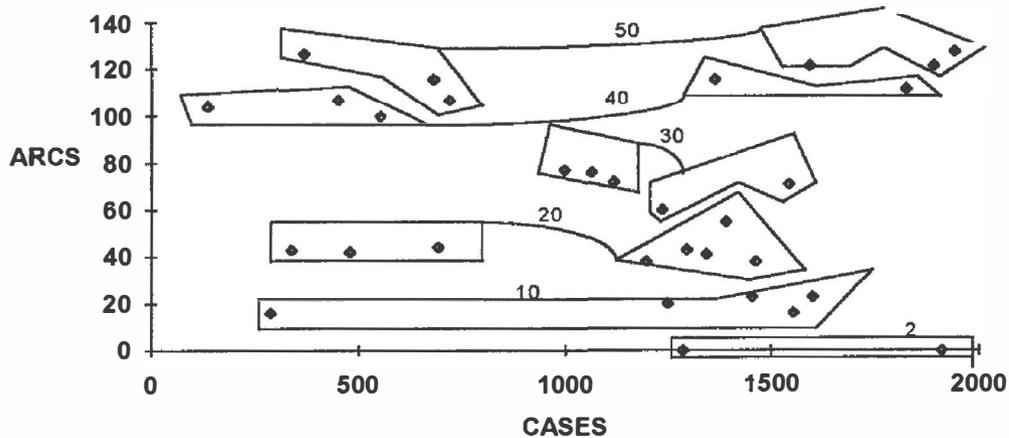

GRAPH I. Relationship between arcs and cases when ordinality is 2. Data points corresponding to BNs with different number of variables are separated into 6 different groups. Numbers for each group denote number of variables.

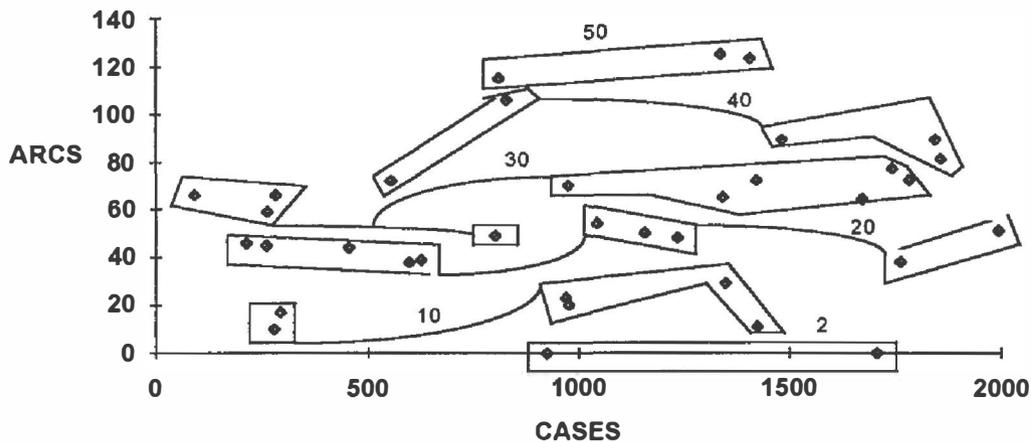

The next step in our analysis is to estimate parameters for the functional relationships we identified. Since the functional form of the relationships appears to be exponential in character, we used the iterative algorithm of SPSS [Norusis 1992] to fit the following models: $M1 = 1 - e^{-C_1 \sqrt{cases}}$ and $M2 = C_2 e^{-C_3 \sqrt{cases}}$. The results of this analysis are given in Table 5. We observe that the explained variability (i.e., fit of the model) which is indicated by $R^2$, is quite good (mean $R^2$ for M1 is 0.6 and for M2 is 0.8, when we model these metrics separately for ordinality of 2 and 3), and thus these models can be used for the assessment of the values of M1, M2 given the sample size we use. Finally, we used our results and K2 to develop a BN model for predicting the expected accuracy of K2, given data attributes. We utilized the following ordering: [number of variables, number of arcs, dimensionality, number of cases, M1, M2]. The BN graph is given in Figure 2, while Appendix I contains the conditional and prior probability distributions.

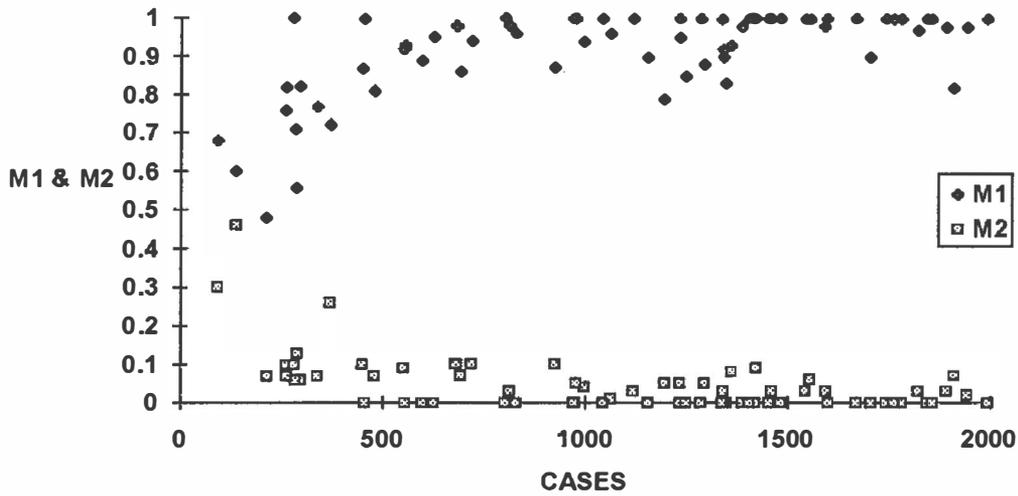

GRAPH III. Relationship between M1 & M2 and the number of cases.

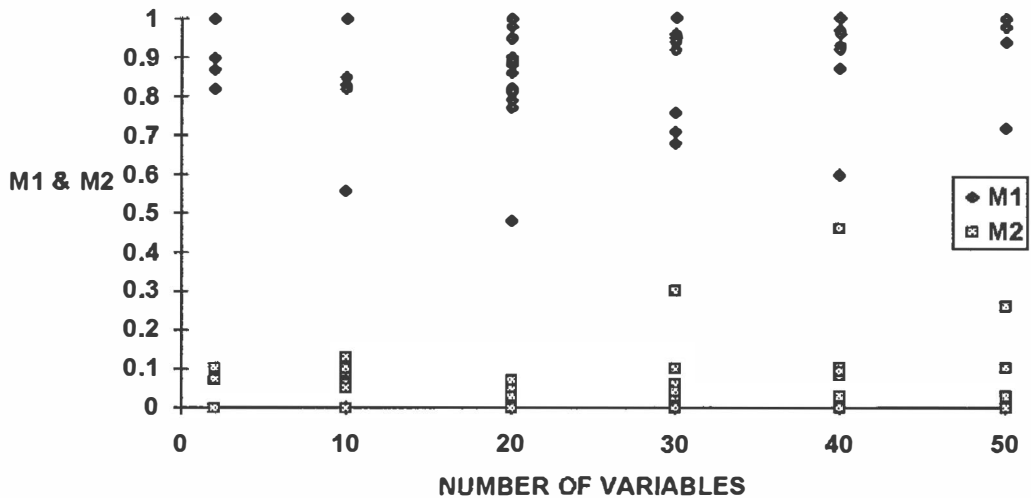

GRAPH IV. Relationship between M1 & M2 and the number of variables.

K2 reveals the fairly complex dependences and independence relationships among the variables, without having any access to domain theory, or the visual/analytical tools we utilized to reach similar conclusions. Using this model (under the assumptions that the underlying data generating process is a BN) we can answer questions of the type: "If the variables are binary and our data set consists of 1200 cases, and we have 20 variables in the model, what is the expected percentage of correct arcs in the model found by K2?" . Or we can ask questions like: "If our data set contains 10 binary variables, how many cases should we have in order for K2 to produce 2% or less extraneous arcs?" We can use any standard BN inference algorithm to answer such questions [Henrion 1990].

## 4 CONCLUSIONS

The results of these experiments are encouraging. Although we used a fairly small number of cases per BN, K2 was able to find the gold standard BN with high accuracy.

Table 5: Non-linear Regression of M1, M2 by Number of Cases

$$M1 = 1 - e^{-C_1 \sqrt{cases}}$$

|  | all cases | ord=2 | ord=3 |
|---|---|---|---|
| $R^2 =$ | 0.57 | 0.65 | 0.56 |
| $C_1$ ($\pm$ SE) = | .09 $\pm$ .004 | .08 $\pm$ .004 | .10 $\pm$ .007 |

$$M2 = C_2\, e^{-C_3 \sqrt{cases}}$$

|  | all cases | ord=2 | ord=3 |
|---|---|---|---|
| $R^2 =$ | 0.58 | 0.78 | 0.79 |
| $C_2$ ($\pm$ SE) = | 1.27 $\pm$ .33 | 1.88 $\pm$ .45 | 2.10 $\pm$ .60 |
| $C_3$ ($\pm$ SE) = | 0.14 $\pm$ .02 | 0.01 $\pm$ .02 | 0.21 $\pm$ .02 |

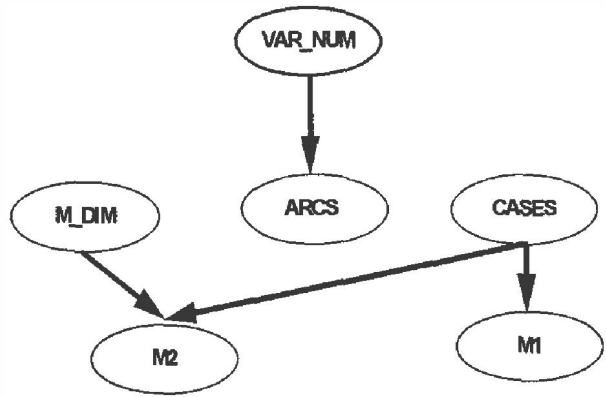

Figure 2: BN Model (Graph Only) of the Variables Relationships

We were also able to identify specific data attributes that determine the expected accuracy of the algorithm, and to build a model for predicting this accuracy. The procedure strongly resembles the process of power and size analysis used in classical statistics, the main difference being that our model was empirically derived. It is important to note that K2 utilizes a simple search method (one step greedy search). In future work we expect to explore the performance of BLN when alternative heuristic search methods are used. Such search methods are likely to diminish or eliminate the need for specification of a total ordering of the variables. The ordering constraint also can be dealt with by using statistical methods similar to those used in the TETRAD II program [Spirtes 1992] to produce (at a first pass) an ordering and then use K2 [Singh 1993].

Other methods for coping with the ordering assumption are to use multiple random orderings and select the one that leads to the most probable BN [Cooper 1992]. Due to the huge number of orderings, this approach would be most practical for BNs with a few variables.

In this experiment we assumed that there were no missing values. Unfortunately in many real-life databases this is not the case. Missing values can be handled normatively as described in [Cooper 1994]. The tractability of this method depends on the domain.

Finally, we parameterized our gold-standard BNs randomly. There is a possibility that BNs that capture real-life processes will deviate from such parameterizations. With our current state of knowledge however, it seems that this is a reasonable initial design choice for an experiment.

### Acknowledgements

We are indebted to Mr. Thanban I. Valappil, and to Dr. Allan R. Sampson for their help with the statistical analysis of the data. Funding was provided by the National Science Foundation under grant # IRI 9111590.

### References

C. Aliferis, E. Chao and G. Cooper "Data Explorer: A Prototype Expert System for Statistical Analysis" Proceedings of the 17th annual SCAMC, 1993, 389-393.

I. Beinlich, H. Suermondt, M. Chavez, G. Cooper "The ALARM monitoring system: A case study with two probabilistic inference techniques for belief networks", Proceedings of the Conference on Artificial Intelligence in Medical Care, London, 1989, 247-256.

G. Cooper, E. Herskovits: "A Bayesian method for the induction of probabilistic networks from data", Machine Learning, 9: 309-347, 1992

G. Cooper: "Current research directions in the development of expert systems based on belief networks", Applied Stochastic Models and Data Analysis, 5: 39-52, 1989.

G. Cooper "A Bayesian method for learning belief networks that contain hidden variables", to appear in: Journal of Intelligent Information Systems, 1994.

R. Fung, S. Crawford "Constructor: A system for the induction of probabilistic models", Proceedings of AAAI 1990, 762-769.

M. Henrion: "An introduction to algorithms for inference in belief networks", In: Uncertainty in

Artificial Intelligence 5, M. Henrion and R. Shachter (Eds.) 1990; 129-138, Amsterdam:North Holland.

M. Henrion: "Propagating uncertainty in Bayesian networks by logic sampling". In: Uncertainty in Artificial Intelligence 2, J. Lemmer and L. Kanal, (Eds.) 1988; 149-163, Amsterdam: North Holland.

E. Herskovits "Computer-Based Probabilistic-Network Construction" Ph.D. thesis, Stanford, 1991.

W. Lam, F. Bacchus "Using causal information and local measures to learn Bayesian networks" Proceedings of Uncertainty in AI 1993, 243-250.

R. Neapolitan: "Probabilistic reasoning in expert systems", New York: John Wiley and Sons 1990.

M. Norusis: "SPSS PC+ vers 4.0 Base manual, Statistics manual, Advanced statistics manual", SPSS Inc 1992.

J. Pearl: "Probabilistic reasoning in intelligent systems", San Mateo, California, Morgan- Kaufmann 1988.

J. Pearl , T. Verma "A statistical semantics for causation" Artificial Intelligence frontiers in statistics: AI and statistics III, Hand DJ (Ed)., New York, Chapman and Hall 1993, p. 327-334.

J. Shavlik, T. Diettrich (Eds.) "Readings in machine learning", San Mateo, California, Morgan-Kaufmann 1990.

M. Singh, M. Valtorta "An algorithm for the construction of Bayesian network structures from data" Proceedings of Uncertainty in AI 1993, 259-265.

P. Spirtes, R. Scheines and C. Glymour "Causation, Prediction and Search", New York, Springer-Verlag 1992.

J. Suzuki "A construction of Bayesian networks from databases based on an MDL principle", Proceedings of Uncertainty in AI 1993, 266-273.

## Appendix I

The following conditional (or prior) probabilities apply to the BN of Figure 2. Note: for each value of the dependent variable we present the conditional probabilities corresponding to the values of the parent variables, created so that the leftmost parent changes values slower, and the rightmost one faster. M_DIM stands for ordinality. Also for the interpretation of the values see Tables 1, 2 and 4.

(a) VAR_NUM: it has no parents

| VAR_NUM value | p(VAR_NUM) |
|---|---|
| 1 | 0.07 |
| 2 | 0.16 |
| 3 | 0.26 |
| 4 | 0.22 |
| 5 | 0.15 |
| 6 | 0.14 |

(b) M_DIM: it has no parents

| M_DIM value | p(M_DIM) |
|---|---|
| 1 | 0.46 |
| 2 | 0.54 |

(c) CASES: it has no parents

| CASES value | p(CASES) |
|---|---|
| 1 | 0.04 |
| 2 | 0.18 |
| 3 | 0.23 |
| 4 | 0.55 |

(d) ARCS: it is determined by VAR_NUM:

| ARCS value | p(ARCS | VAR_NUM) |
|---|---|
| 1 | 0.63  0.53  0.05  0.05  0.07  0.08 |
| 2 | 0.13  0.33  0.86  0.21  0.07  0.08 |
| 3 | 0.11  0.07  0.05  0.68  0.43  0.08 |
| 4 | 0.13  0.07  0.05  0.05  0.43  0.77 |

(e) M1: it is determined by CASES:

| M1 value | p(M1 | CASES) |
|---|---|
| 1 | 0.13  0.11  0.05  0.02 |
| 2 | 0.38  0.11  0.05  0.02 |
| 3 | 0.13  0.50  0.19  0.20 |
| 4 | 0.10  0.06  0.29  0.09 |
| 5 | 0.13  0.06  0.19  0.16 |
| 6 | 0.13  0.17  0.24  0.50 |

(f) M2: it is determined by M_DIM and CASES:

| M2 value | p(M2 | M_DIM, CASES) |
|---|---|
| 1 | 0.17  0.17  0.10  0.17  0.10  0.53  0.36  0.74 |
| 2 | 0.17  0.17  0.10  0.08  0.20  0.20  0.40  0.09 |
| 3 | 0.17  0.17  0.40  0.58  0.50  0.13  0.16  0.09 |
| 4 | 0.17  0.33  0.30  0.08  0.10  0.07  0.04  0.04 |
| 5 | 0.33  0.17  0.10  0.08  0.10  0.07  0.04  0.04 |